\documentclass[10pt,twocolumn,letterpaper]{article}

\usepackage{iccv}
\usepackage{times}
\usepackage{epsfig}
\usepackage{graphicx}
\usepackage{amsmath}
\usepackage{amssymb}
\usepackage{mathtools}
\usepackage{bbm}
\usepackage{makecell, verbatim, sidecap}
\usepackage[breaklinks=true,bookmarks=false]{hyperref}

\iccvfinalcopy %

\usepackage{color}

\def\block1{{$\sf block1$}}
\def\block2{{$\sf block2$}}
\def\block3{{$\sf block3$}}
\def\block4{{$\sf block4$}}

\usepackage[margin=4pt,font=small,labelfont=bf,labelsep=endash,tableposition=top]{caption}

\renewcommand{\texttt}[1]{ $ {{\tt #1} } $}

\usepackage{xspace}

\renewcommand{\vec}[1]{\ensuremath{\pmb{#1}}}
\newcommand{\mat}[1]{\ensuremath{\mathbf{#1}}}
\newcommand{\set}[1]{\ensuremath{\mathscr{#1}}}

\makeatletter
\@tfor\next:=abcdefghijklmnopqrstuvwxyxz\do
 {\begingroup\edef\x{\endgroup
    \noexpand\@namedef{v\next}{\noexpand\vec{\next}}%
  }\x}
\@tfor\next:=ABCDEFGHIJKLMNOPQRSTUVWXYZ\do
 {\begingroup\edef\x{\endgroup
    \noexpand\@namedef{m\next}{\noexpand\mat{\next}}%
  }\x}
\@tfor\next:=ABCDEFGHIJKLMNOPQRSTUVWXYZ\do
 {\begingroup\edef\x{\endgroup
    \noexpand\@namedef{s\next}{\noexpand\set{\next}}%
  }\x}
\makeatother

\usepackage{marvosym}

\begin{document}

\title{AGSFCOS:Based on attention mechanism and Scale-Equalizing pyramid network of object detection}


\author{
	Li Wang ~ ~ ~ ~ ~ 
	Wei Xiang\textsuperscript{\Letter}\thanks{Corresponding author:21500068@swun.edu.cn} ~ ~ ~ ~ ~
	Ruhui Xue ~ ~ ~ ~ ~ 
	Kaida Zou ~ ~ ~ ~ ~ 
	Laili Zhu ~ ~ ~ ~ ~	\\ \\
Key Laboratory of Electronic and Information Engineering, State Ethnic Affairs Commison, \\Southewst Minzu University,Chengdu City,Sichuan Province
}
\maketitle

\begin{abstract}
	Recently, the anchor-free object detection model has shown great potential for accuracy and speed to exceed anchor-based object detection. Therefore, two issues are mainly studied in this article: (1) How to let the backbone network in the anchor-free object detection model learn feature extraction? (2) How to make better use of the feature pyramid network?In order to solve the above problems, Experiments show that our model has a certain improvement in accuracy compared with the current popular detection models on the COCO dataset, the designed attention mechanism module can capture contextual information well, improve detection accuracy, and use sepc network to help balance abstract and detailed information, and reduce the problem of semantic gap in the feature pyramid network.Whether it is anchor-based network model YOLOv3, Faster RCNN, or anchor-free network model Foveabox, FSAF, FCOS. Our optimal model can get 39.5\% COCO AP under the background of ResNet50.


\end{abstract}

\begin{figure*}[htb]
	\begin{center}
		\includegraphics[width=.9\linewidth]{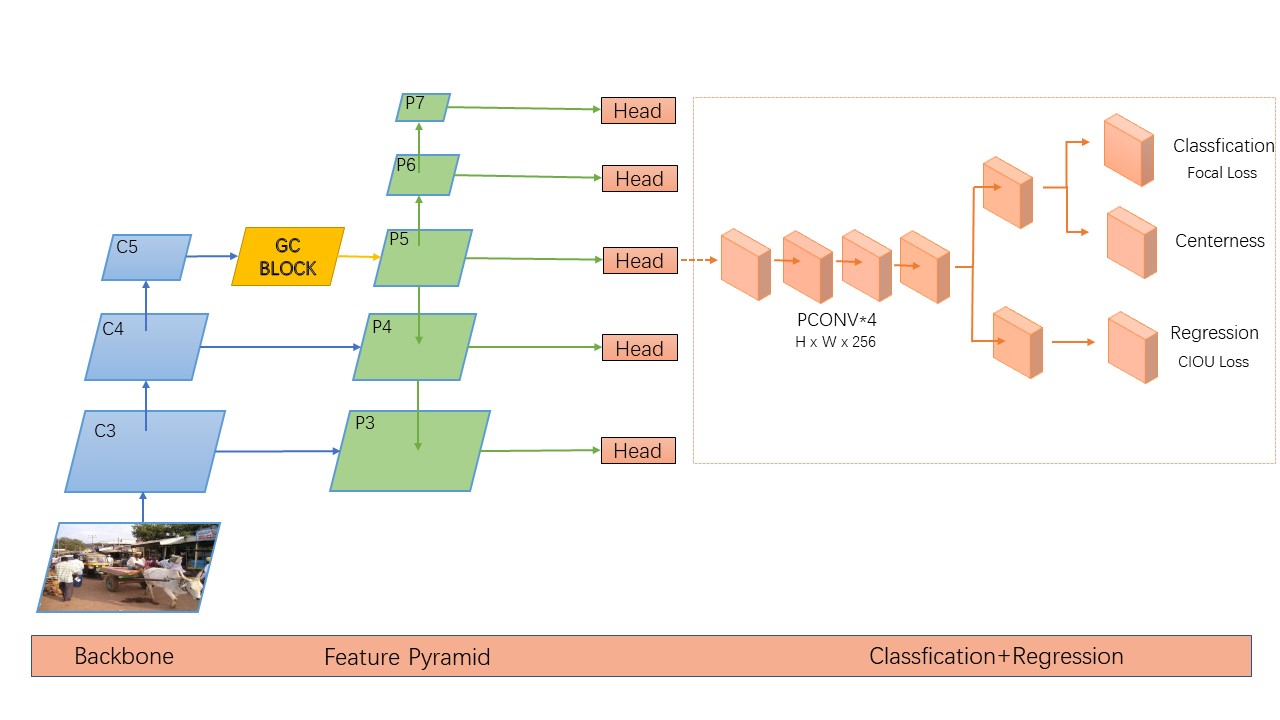}
	\end{center}
	\caption{AGSFCOS network structure
	}
	\label{fig:figure1}
\end{figure*}
\section{Introduction}

With the rapid development of deep learning, object detection is a basic computer vision task that com-bines the two tasks of object location and recognition. On the one hand, the effect of object detection is re-lated to the research of many high-level tasks such as object tracking and pedestrian re-recognition, which has strong theoretical value; on the other hand, ob-ject detection has strong application value in the con-text of transportation, security, medical treatment, etc., so it has always been one of the important re-search tasks of computer vision. Most of these algo-rithms use anchors to solve the scale and scale changes of the target. The main purpose of the an-chor is to determine whether there is an object of in-terest in the reference frame and determine the posi-tion of the object from the reference frame. First set an area for the anchor, and then change the aspect ratio of the anchor, the network model will predict the type and offset of the designed anchor, and perform regression through the offset to obtain the best pre-dicted frame, such as Fast RCNN\cite{article1}, Faster RCNN\cite{article2}, etc., RoI Pooling proposed by Fast RCNN accelerates the entire network process, unifies the category out-put task and candidate box regression task, and combines the advantages of Faster RCNN and YO-LO\cite{article3}, SSD\cite{article4} can combine high-level and low-level feature maps, and use multi-scale ideas to regress the extracted features.
Although the anchor-based model has good detection effect, there are still many problems: (1) Anchor relies on pre-setting, which generally needs to be set manually, and then obtain better results through a large number of experiments. (2) When setting the anchor, it’s unable to take into account the very large and very small targets, which easily makes it difficult for the model to handle objects with large scale changes. (3) The anchor-based object detection model will generate a corresponding anchor box on each feature map, the number of candidate frames generated increases,the problem of imbalance between positive and negative samples, and the amount of calculation is also increased.
In recent years, the anchor-free object detection model has also received a lot of attention. They do not need to design an anchor. Compared with the traditional anchor-based object detection model, it has the following advantages: (1) No need to manually set the anchor, reducing the amount of calculation. (2) The network structure framework is simpler. (3) Better feature selection can be achieved.
In this article, we introduce a new anchor-free object detection model AGSFCOS. The designed attention mechanism module captures global information to obtain larger receptive fields and contextual information which can help improve detection accuracy. The main function is to focus on the object area and reduce the interference of the image background and negative sample information, especially when the background is complex. The designed scale-balanced pyramid helps to balance abstract and detailed information, and the stacked pyramid convolution directly extracts spatial and scale features, reducing the problem of large semantic differences in the feature pyramid network. Our method is improved on the FCOS model.


\section{Related Work}
The object detection model based on deep learning has been developed so far and can be divided into two branches: anchor-free and anchor-based.
Anchor-based detectors. The current detection models can be divided into two categories: one-stage and two-stage detection. One-stage detection includes YOLOv2\cite{article5}, YOLOv3\cite{article6}, RetinaNet\cite{article7}, SSD, DSOD\cite{article8}, etc. Single-stage object detection is defined as a one-step completion. the algorithm does not need to generate a candidate frame, only needs to map the preset candidate frame to the feature map, the category probability and position coordinates of the object can be marked directly to locate the object frame, and transformed the problem into a regression problem. After a single test, the final test result can be directly obtained, so the speed tends to be faster. Two-stage detection includes R-FCN\cite{article9}, FPN\cite{article10}, Cascade R-CNN\cite{article11}, Libra R-CNN\cite{article12}, TridentNet\cite{article13}, etc. The two-stage object detection algorithm defines the detection frame as a process from coarse to fine. The candidate frames are obtained through selective search and Region Proposal Network algorithms, and then the images in these candidate frames are used as input to judge the object category, and the positioning of the bounding box of the object position completed by regression, so the algorithm has better detection accuracy and precision.
Anchor-free detectors. Although anchor-based detectors currently dominate, anchor-free detectors are still evolving. Recently, Anchor-free object detection algorithms have received a lot of attention, and they do not rely on pre-defined anchor frames. Compared with the traditional anchor-based method, anchor-free detectors don’t have artificial hyperparameters for the anchor frame configuration. The detection structure is usually simpler and has the potential to surpass the anchor-based method in terms of speed and accuracy. The object detection model based on anchor-free is mainly divided into a object detection model based on key points and a target detection model based on the central region. Early work such as DenseBox\cite{article14}, and UnitBox\cite{article15} explored another direction of region proposal, and was used in scene text detection and pedestrian detection. Recent work has promoted the performance of anchor-free detectors, gradually surpass some anchor-based detectors. Most of these anchor-free detectors are one-stage. For example, CornerNet\cite{article16} ,CornerNet-Lite\cite{article17}, ExtremeNet\cite{article18} and CenterNet\cite{article19} convert object detection into corner points for detection. FSAF\cite{article20}, Guided Anchoring\cite{article21}, FCOS\cite{article22} and FoveaBox\cite{article23} encode bbox into anchor points and decode into distance from point to boundary.

\section{Our network structure}
In this section, we mainly describe the AGSFCOS ob-ject detection network model in detail. The AGFCOS network model mainly includes extract feature net-work, attention mechanism GC block, feature pyra-mid network, scale equalization pyramid network, and classification and regression modules. The basic structure of the network model is shown in Figure~\ref{fig:figure1}.

\subsection{Feature extraction network}

\begin{figure}[htb]
	\flushleft
	\includegraphics[width=1.37\linewidth]{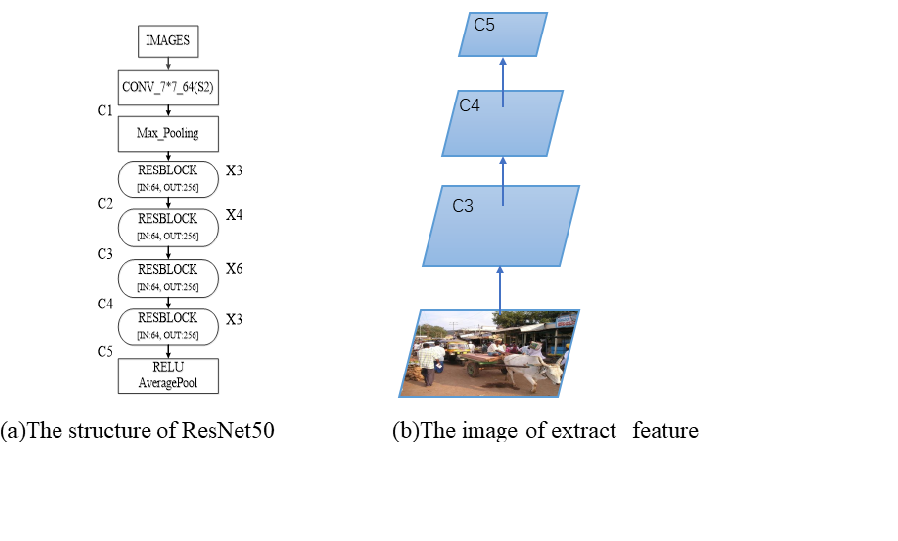}
	\caption{the structure of ResNet50}
	\label{fig:figure2}
\end{figure}

This paper does not involve the improvement of the feature extraction network, and chooses a model with a smaller number of layers can effectively re-duce the training cost and test time, so the feature ex-traction network in this paper uses ResNet50 uni-formly. The network structure of ResNet50 is shown in Fig2, but this article only uses the features of the C3, C4, and C5 layers in the ResNet50 network struc-ture, and then performs the features of the C3, C4, and C5 layers respectively (8, 16, 32) The down-sampling is shown in Figure~\ref{fig:figure2}

\subsection{Attention mechanism-GC block}
The features extracted from the feature extraction network are put into this module, which includes three sub-modules: context modeling, transformation features, and feature fusion. The attention mechanism module of GC block\cite{article24} is shown in the Fig4, which is mainly divided into three steps:
\begin{itemize}
\item[(1)] First, input x is convoluted by 1 * 1 to get a value of 1 * h * W, and then the weight of attention can be obtained by passing it through the Softmax layer.
\item[(2)] Then global attention pooling is used to model the context of the input features. This step allows the entire input to cover the receptive field to capture the dependency between channels.
\item[(3)] Finally, the feature elements obtained from the above two steps are added element by element to complete the feature fusion.
\end{itemize}

\begin{figure}[htb]
	\flushleft
	\includegraphics[width=1.56\linewidth]{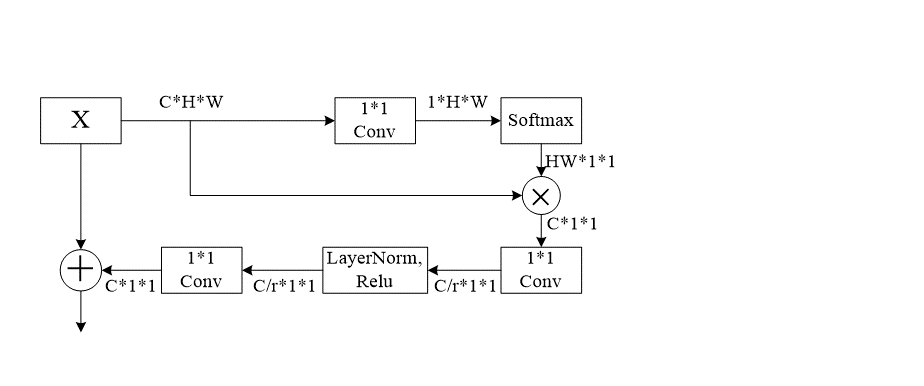}
	\caption{GC block structure}
	\label{fig:figure3}
\end{figure}

The specific expression is shown in Equation(1):
\begin{equation}
	\begin{aligned}
		z_i=x_i+W_{v_2}ReLU(LN(W_{v_1}\sum_{j=1}^{N_p}\frac{e^{w_k}x_j}{\sum_{m=1}^{N_p}e^{w_k}x_m}x_j))
	\end{aligned}
\end{equation}
Among them,
$\frac{e^{w_k}x_j}{\sum_{m=1}^{N_p}e^{w_k}x_m}$
is the weight of global attention pooling, and
$W_{v_2}ReLU(LN(W_{v_1}))$
represents the transformation of the bottom network structure. i is the index of the query location, j lists all the locations,$W_{v_1}$and $W_{v_2}$represent the linear transformation matrix.
GC block models the context of the whole input fea-ture, so the whole input feature can cover the recep-tive field, which is more capable of supplementing contextual information than the local receptive field of the ordinary convolutional layer. In addition, the network only extracts features through convolutional stacking, which will lead to a lack of diversity in the network extracted features, and the GC block com-posed of NLNet \cite{article25} and SENet \cite{article26} just compensates for the lack of diversity.

\subsection{Feature Pyramid Network}
Put the features obtained after using the GC block in-to the feature pyramid network. The feature pyramid is not used alone in the object detection model. It needs to be used together with the feature extraction backbone network, as shown in Figure~\ref{fig:figure5}.
Figure~\ref{fig:figure5} shows the feature pyramid network of FCOS network model, the backbone uses ResNet50 to ex-tract features. There are many down sampling stages in this part. Each down sampling operation will dou-ble the feature map on the original basis, so as to get many feature maps of different sizes. In AGSFCOS network model, only 1/8, 1/16, and 1/32 feature maps are used. From the leftmost side of the figure, you can see the feature maps of the input image, which are C3, C4 and C5. After a 1 * 1 convolution, the top-level feature map C5 can get P5, and then C4 and P5 carry out feature fusion. but it can be seen from the figure that the size of C4 is twice that of P5, so we need to double P5 to get P4, and then combine with C4.

\begin{figure}[htb]
	\centering
	\includegraphics[width=1.4\linewidth]{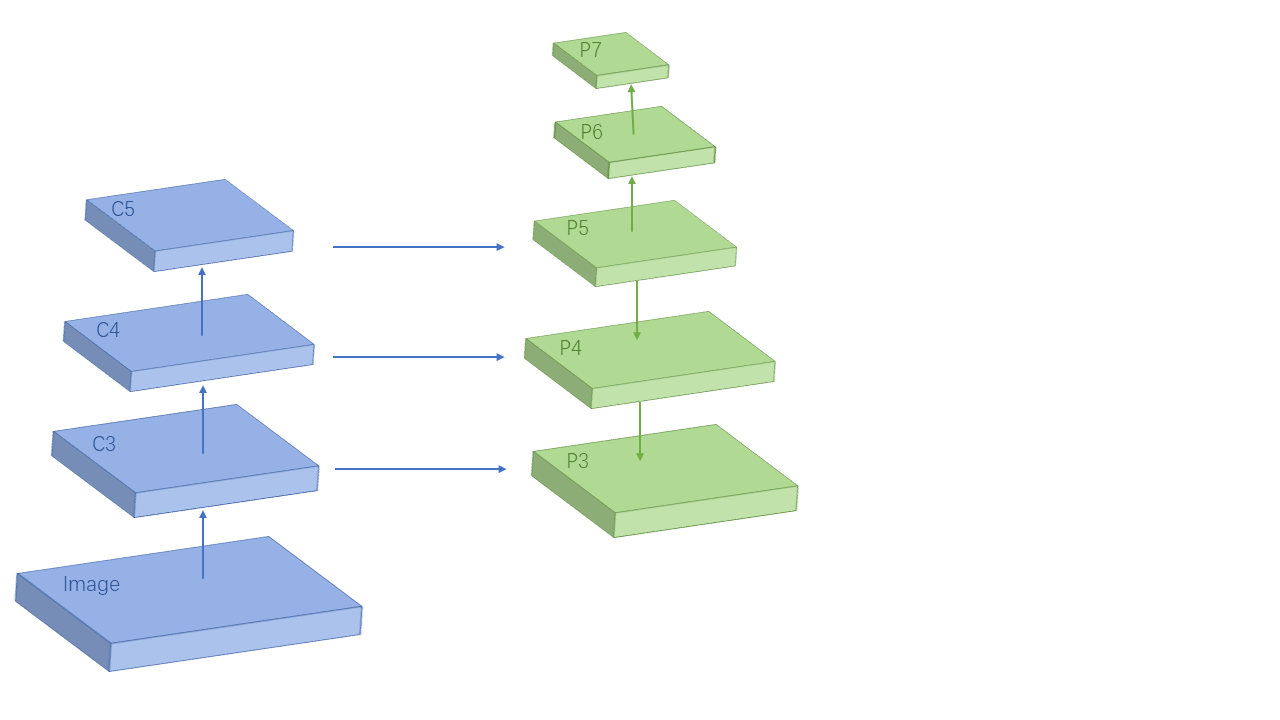}
	\caption{The feature pyramid network structure of AGSFCOS}
	\label{fig:figure4}
\end{figure}

%


\subsection{Scale-Equalizing Pyramid Network-SEPC}
If only the feature pyramid network is used to complete the multi-scale strategy, there will be a large semantic gap in each layer, so in this paper, the scale-balanced pyramid network is used to alleviate such a problem \cite{article27}.
Pyramid convolution performs 3D convolution in two dimensions of scale and space, but if each point in the figure is regarded as a feature point, then pyramid convolution can be considered as multiple different 2D convolution kernels. However, in the pyramid network, the scale of each layer is different. With the increase of the number of layers in pyramid network, the scale of the feature map will become smaller. Pyramid convolution sets different step sizes for multiple different convolution kernels to solve the problem of scale mismatch. For pyramid convolution with N = 3, the step size of the first kernel is 2, and the step size of the last kernel is 0.5. Then the output of pyramid convolution can be expressed as c (2):

\begin{equation}
	\begin{aligned}
		y^l=w_1*s0.5x^{l+1}+w_0*x^l+w_{-1}*s2x^{l-1}
	\end{aligned}
\end{equation}

Among them, represents a certain layer of the pyramid network, is three independent 2Dl convolution kernels, x is the input feature map, s2 represents the convolution with a step size of 2, s0.5 represents the convolution with a step size of 0.5, and then through an ordinary convolution with a step size of 1 plus a bilinear up sampling layer, it is realized as Equation (3):
\begin{equation}
	\begin{aligned}
		y^l=uspsample(w_1*x^{l+1})+w_0*x^l+w_1*s2x^{l-1}
	\end{aligned}
\end{equation}

\begin{figure}[htb]
	\centering
	\includegraphics[width=\linewidth]{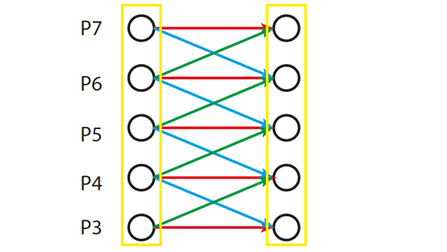}
	\caption{The structure of PCONV}
	\label{fig:figure5}
\end{figure}

\begin{figure}[htb]
	\centering
	\includegraphics[width=1.2\linewidth]{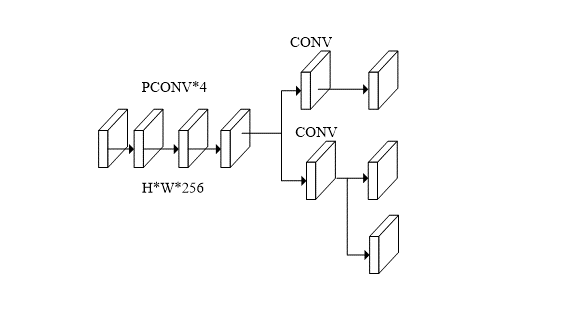}
	\caption{AGSFCOS head work}
	\label{fig:figure6}
\end{figure}

Similar to traditional convolution, pyramid convolution also has zero padding. For the bottom layer ( ) of the pyramid, the last term of Equa-tion (3) is invalid; for the top layer of the pyramid, the first term of Equation (3) is invalid. Although three convolution operations are performed on each layer of the pyramid, the calculation amount of the pyramid convolution is only 1.5 times that of the ordinary convolution.
Four consecutive convolutions can be replaced by pyramid convolutions with three scales. In the deep neural network, the stacked pyramid convolution corresponds to the stacked convolution module, so it can bring less calculation. But each pyramid convolution still brings additional calculations. As an alternative, four pyramid convolution, classification and regression parts form the shared parameters of the head network. In order to obtain the results of both classification and regression, only 3*3 convolutions need to be added after four pyramid convolutions. Figure~\ref{fig:figure5} is the structure diagram of PCONV, Figure~\ref{fig:figure6} is the head network of the AGSFCOS network model designed by PCONV. The head network using PCONV only needs to go through three pyramid convolutions and the structure is simpler.

\subsection{Classification and regression}
Three loss functions were used in training the AGSFCOS model, including the loss function focal loss for classification \cite{article28}, CIOU loss for regression \cite{article29} and the binary cross entropy loss of the center-ness strategy (Binary Cross Entropy loss, BCE loss).

\subsubsection{Classification}
\begin{figure}[htb]
	\centering
	\includegraphics[width=1.14\linewidth]{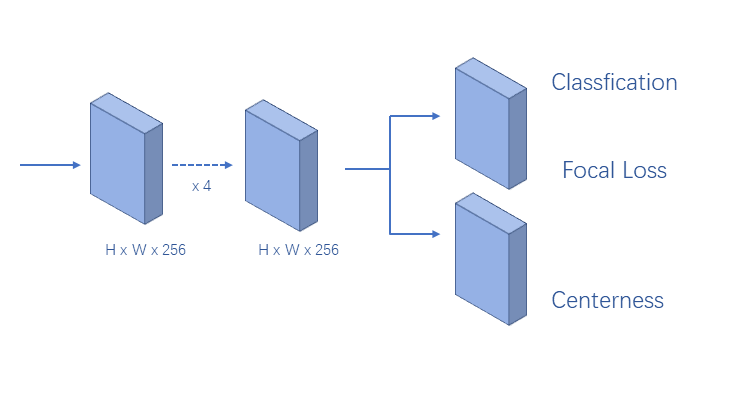}
	\caption{The image of Classification structure}
	\label{fig:figure7}
\end{figure}

The classification part of the AGSFCOS model is mainly divided into classification and center-ness. First, the features extracted by the multi-scale network are subjected to four 3*3 convolutions, and then the Focal Loss and the binary cross-entropy loss function are used to Complete classification. As shown in Figure~\ref{fig:figure7}.

Using C two classifications to output C predicted values, C represents the number of categories of the object, classification is to make binary classifications of each category, and the loss function used is Focal Loss, as shown in Equation (4).
\begin{equation}
	\begin{aligned}
		L_{fl} =
		\begin{cases}
			-\alpha(1-y^\prime)^\gamma\log{y^\prime},  & y=1 \\
			-(1-\alpha)y^{\prime\gamma}\log(1-y^\prime), & y=0
		\end{cases}
	\end{aligned}
\end{equation}
 $y^\prime$is the output of the activation function, with a value between 0 and 1. It can reduce the loss of easy to classify samples, and focus on more easy to misclassify and difficult samples. In addition, the balance factor can balance the positive and negative samples. In this paper, Center-ness is used to calculate the distance between each pixel and the object center point, which can be used to reduce the predicted points that are far away from the object center point. Center-ness will add a branch to each prediction layer, which is parallel to the classification. Center-ness can ensure that the predicted bounding box is as close to the center as possible, as shown in Equation (5).

 \begin{equation}
 	\begin{aligned}
 		centerness = \sqrt{{\frac{min(l,r)}{max(l,r)}}×{\frac{min(t,b)}{max(t,b)}}}
 	\end{aligned}
 \end{equation}
Among them are the four predicted values of each pixel from the bounding box.

\subsubsection{Regression}
In this paper, we used CIOU Loss. CIOU Loss not only considers the overlapping area, but also considers the distance between the center points and the aspect ratio. Compared with GIOU Loss and IOU Loss, CIOU Loss can achieve better convergence and accuracy. CIOU Loss introduces a penalty term for the aspect ratio of the frame, which takes into account the aspect ratio of the frame. The definition is shown in Equation (6):
\begin{equation}
	\begin{aligned}
		L_{IoU} = 1-IOU+\frac{\rho^2(b,b^{gt})}{c^2}+\alpha\upsilon
	\end{aligned}
\end{equation}
Among them,$\upsilon=\frac{4}{\pi^2}(arctan\frac{\omega^{gt}}{h^{gt}}-arctan\frac{\omega}{h})^2$,$\alpha=\frac{\upsilon}{(1-IoU)+\upsilon}$
represent the Euclidean distance between two center points. c represents the diagonal distance of the smallest closed area that can contain both the prediction box and the ground truth box.
\begin{table*}[!tbh]
	\centering
	\caption{Comparative experiment using GIOU LOSS and CIOU LOSS}
	\label{table:table1}
	\begin{tabular}{ |c |c|c|c|c|c|c|c|c|c|c|c|c}
		\hline
		Method & AP & AP50 & AP75 & APS & APM & APL & AR & ARS & ARM & ARL \\	
		\hline
		FCOS & 36.4 & 54.5 &38.9 &20.2 & 39.7 & 47.4 & 53.2 & 33.5 &57.7 & 67.8 \\
		\hline
		+GC & 37.1 &56.0 &39.5 & 21.3 & 40.5 & 47.9 & 53.5 & 33.7 & 57.9 & 68.8\\
		\hline									
		+GC+CIOU & 37.5 & 56.3& 39.8 & 21.8 & 41.0 & 49.1 & 53.7 & 33.8 & 58.2 & 68.8  \\
		\hline									
	\end{tabular}
	%
	
\end{table*}


\begin{table*}[!tbh]

	\centering
	\caption{Comparative experiments using GC BLOCK}
	\label{table:table2}
	\begin{tabular}{ |c |c|c|c|c|c|c|c|c|c|c|c|c}
		\hline
		Method & AP & AP50 & AP75 & APS & APM & APL & AR & ARS & ARM & ARL \\
		\hline
		FCOS & 36.4 & 54.5 &38.9 &20.2 & 39.7 & 47.4 & 53.2 & 33.5 &57.7 & 67.8 \\
		\hline
		+GC & 37.1 &56.0 &39.5 & 21.3 & 40.5 & 47.9 & 53.5 & 33.7 & 57.9 & 68.8\\
		\hline									
		+GC+CIOU & 37.5 & 56.3& 39.8 & 21.8 & 41.0 & 49.1 & 53.7 & 33.8 & 58.2 & 68.8  \\
		\hline									
		
	\end{tabular}
	%
	
\end{table*}

\begin{table*}[!tbh]
	\centering
	\caption{Comparative experiments using SEPC}
	\label{table:table3}
	\begin{tabular}{ |c |c|c|c|c|c|c|c|c|c|c|c|c}
		\hline
		Method & AP & AP50 & AP75 & APS & APM & APL & AR & ARS & ARM & ARL \\
		\hline
		FCOS & 36.4 & 54.5 &38.9 &20.2 & 39.7 & 47.4 & 53.2 & 33.5 &57.7 & 67.8 \\
		\hline
		+SEPC & 39.5&57.1& 42.4&22.3&	43.2&	52.1&	55.5&	34.3&	60.6&	71.3\\									
		\hline									
		
	\end{tabular}
	%
	
\end{table*}

\begin{table*}[!tbh]
	\centering
	\caption{AGSFOS vs. other state-of-the-art detectors}
	\label{table:table4}
	\begin{tabular}{ |c |c|c|c|c|c|c|c|c|c|c|c|c}
		\hline
		Method & Anchor-free & backbone&AP&AP50 & AP75 & APS & APM & APL \\
		\hline
		Faster RCNN & No & ResNet50 &36.2&	59.1&	39.0&	18.2&	39.0&	48.2\\
		YOLOv2 & No &Darknet-19 &21.6&	44.0&	19.2&	5.0&22.4&	35.5\\									
		YOLOv3 & No & Darknet-53& 27.9&	49.2&	28.3&	10.5&	30.1&	43.8 \\
		SSD & No & ResNet101& 31.2&	50.4&	33.3&	10.2&	34.5&	49.8 \\
		Retinanet & No & ResNet50&36.3&	55.3&	38.6&	19.3&	40.0&	48.8 \\
		Foveabox & Yes & ResNet50& 35.2&	54.0&	38.4&	20.3&	39.7&	47.5 \\
		FSAF & Yes & ResNet50& 36.0&	55.5&	37.7&	19.6&	39.6&	48.2 \\
		FCOS & Yes & ResNet50& 36.4&	54.5&	38.9&	20.2&	39.7&	47.4 \\
		AGSFCOS & Yes & ResNet50& 39.5&	57.1&	42.4&	22.3&	43.2&	52.1\\
		\hline									
		
	\end{tabular}
	%
	
\end{table*}

\section{Experiments}
AGSFCOS model experiment is completed on the mmdetection framework of the deep learning object detection toolbox based on pytorch. We use the COCO dataset to train the model and compare it with other models, the validation set has 5k images, and the test set has 123k images. In order to ensure a fair experimental comparison, all of our experiments are completed on a NVIDIA GeForce RTX 2080Ti.
Training parameters: We use ResNet50 as the feature extraction network in this paper, batch size is set to 4, stochastic gradient descent (SGD) is used to train 12 epochs, the initial learning rate is set to 0.001, and it is reduced to one tenth of the original after the 8th and 11th epochs. The weight decay rate is set to 0.0001, and the momentum is 0.9, the image size is set to (800,1333).
Evaluation indicators: The main performance evaluation indicators used in this article are average Precision, (AP), average Recall rate (AR). AP50 and AP75 represent the mAP value when the IOU threshold is 0.5 and 0.75, respectively. APS, APM, and APL represent the average accuracy of small objects, medium objects, and large objects. ARS, ARM, and ARL respectively represent the average recall rate of small objects, medium objects, and large objects.

\subsection{Comparison of different regression loss}
As we mentioned before, when the model is regressing, we use CIOU LOSS instead of GIOU LOSS, which can achieve better convergence and accuracy. As can be seen from Table 1, FCOS model using CIOU LOSS can Increase the AP value from 36.4\% to 36.7\%.

\subsection{Comparison with attention mecha-nism GC block}

After the feature extraction network ResNet50, we use the attention mechanism module GC block, as can be seen from Table 2, After adding the GC block, the AP value increased by 0.7\%, and the values of AR, ARS, ARM and ARL added to the CIOU LOSS model increased by 0.5\%, 0.3\%, 0.5\%, and 1\% re-spectively. From the perspective of the improvement effect, the average recall rate of the improved model is still significantly improved, indicating that the  im-provement of the loss function and the addition of the attention mechanism module can effectively increase the recall rate.

\subsection{With or without SEPC}
The deformable convolution is used to adapt to the corresponding irregularities between the actual fea-tures, and to maintain a balanced scale. The adapta-bility of the deformable convolution can handle the large inter-layer blurring of the feature pyramid. It can be seen from Table 3 that the AGSFCOS model using the SEPC module has a significant improve-ment in AP regardless of whether it is a middle, large or small target. ARS, ARM, and ARL have also in-creased by 1.2\%, 0.9\%, and 3.5\%. The overall de-scription of the effectiveness of the AGSFCOS model.

\subsection{Comparison with State-of-the-art Detectors}

We compared FCOS with other latest object detectors in the MS-COCO benchmark. For these experiments. As shown in Table 4, with ResNet50-GC-SEPC, the performance of our FCOS on RetinaNet with the same backbone network is 3.2\% higher than in AP. Whether compared with anchor-based network mod-els such as the one-stage model YOLO series or the two-stage model Faster RCNN, the detection results are still improved by 13\% and 3.3\%. Under the same backbone, compared with anchor-free object detec-tion algorithms such as FSAF and Foveabox, the AP value of AGSFCOS still exceeds them. From the ex-perimental results, it can be seen that the improve-ment effect of AGSFCOS on the average accuracy of large, medium and small targets is very obvious.

\begin{figure}[htb]
	\flushleft
	\includegraphics[width=1.15\linewidth]{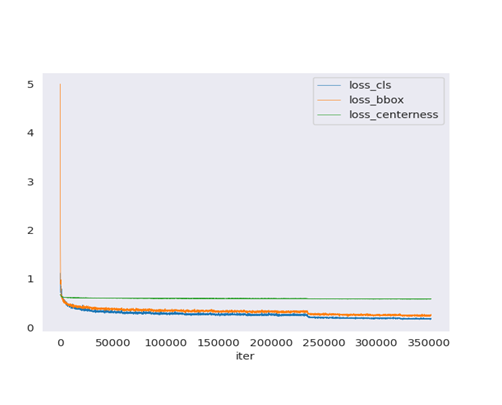}
	\caption{Loss function of AGSFCOS model training}
	\label{fig:figure8}
\end{figure}

\subsection{The Loss and AP for AGSFCOS}

\begin{figure}[htb]
	\flushleft
	\includegraphics[width=1.1\linewidth]{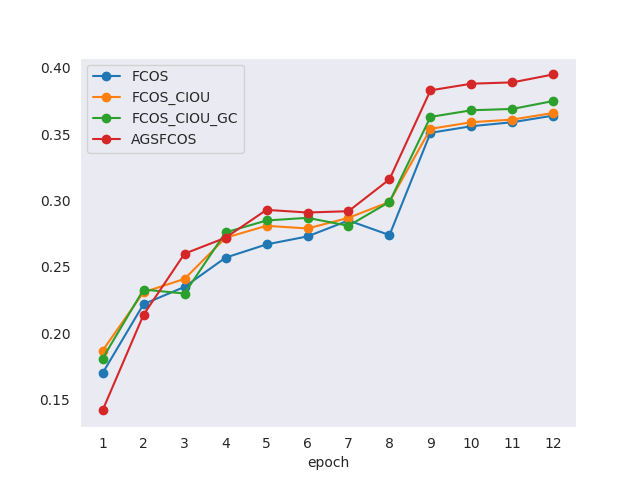}
	\caption{Comparison of AP value in AGSFCOS, FCOS,FCOS\_CIOU, FCOS\_CIOU\_GC}
	\label{fig:figure9}
\end{figure}

In Figure\ref{fig:figure8}, the horizontal axis represents the number of iterations of AGSFCOS network model training, and the vertical axis represents the change in the loss value in the training process. The green line is the loss function change curve of Center-ness in the training of the AGSFCOS model, and the orange line is the regression. CIOU Loss function change curve, the blue line is the loss function change curve of Focal Loss. It can be seen from the figure that the regression loss function and classification loss function of the FCOS network model have a higher loss at the beginning, but after training 250,000 times, the curve stabilizes, and then the learning rate is reduced, and you can see that the loss curve has dropped significantly and then further tends to be smooth, and finally stop training at 350000 times.
In Figure~\ref{fig:figure9}, the horizontal axis represents the training epoch, the vertical axis represents the change in the mAP value of the network model at epoch[0,12] at the end of each training, and the blue is the mAP value of the FCOS network model at epoch[0 ,12], the orange line is the change of the mAP value of the FCOS\_CIOU network model on epoch[0,12], and the green line is the change of the mAP value of the FCOS\_CIOU\_GC network model on epoch[0,12]. The red line is the change of the AGSFCOS model on epoch[0,12]. It can be seen from Figure ~\ref{fig:figure9} that after epoch is 4, AGSFCOS far exceeds the mAP value of other models.

\section{Conclusion}
We propose an object detection model AGSFCOS based on anchor-free. As shown in the experiment, this model not only has advantages over anchor-based target detection models RetinaNet, Faster RCNN, YOLO and SSD, but also has advantages over anchor-free target detection models FSAF and Foveabox. In comparison, the AP values has also been improved. In view of its effectiveness and accuracy, we hope that the AGSFCOS model can effectively solve the alternative of anchor-based object detection. We also believe that the AGSFCOS model can be extended to solve other tasks.



%

\section*{Acknowledgment}
This work is financially supported by the Fundamental Research Funds for Central University, Southwest Minzu University ( 2020NYBPY02).



\begin{thebibliography}{30}
\bibitem{article1}Girshick R. Fast R-CNN. IEEE International Conference on Computer Vision (ICCV), 2015:1440-1448. https://doi.org/10.1109/ICCV.2015.169.
\bibitem{article2}Ren S, He K, Girshick R, et al. Faster R-CNN: Towards Real-Time Object Detection with Region Proposal Networks. IEEE Transactions on Pattern Analysis and Machine Intelligence, 2015, 39(6):1137-1149. https://doi.org/10.1109/TPAMI.2016.2577031.
\bibitem{article3}Joseph R, Santosh D, Ross G. You only look once: Unified, real-time object detection. IEEE Conference on Computer Vision and Pattern Recognition (CVPR). 2016:779-788.https://doi.org/10.1109/CVPR.2016.91.
\bibitem{article4}Liu W, Anguelov D, Erhan D, et al. Ssd: Single shot multibox detector. Proceedings of the European Conference on Computer Vision, 2016:21-37. https://doi.org/10.1007/978-3-319-46448-0\_2
\bibitem{article5}Redmon J, Farhadi A. YOLO9000: Better, Faster, Stronger. IEEE Conference on Computer Vision \& Pattern Recognition. IEEE,
2017:6517-6525. https://doi.org/10.1109/CVPR.2017.690
\bibitem{article6}Redmon J, Farhadi A. YOLOv3: An Incremental Improvement. arXiv e-prints, 2018.
\bibitem{article7}Tsung-Yi Lin, Priya Goyal, Ross Girshick, Kaiming He, andPiotr Doll'ar. Focal loss for dense object dete-ction. Inproc.IEEE
Conf. Comp. Vis. Patt. Recogn.2017:2980-2988. https://doi.org/10.1109/TPAMI.2018.2858826.
\bibitem{article8}Shen Z, Zhuang L, Li J, et al. DSOD: Learning Deeply Supervised Object Detectors from Scratch. 2017 IEEE International Conference on Computer Vision (ICCV). IEEE, 2017. https://doi.org/10.1109/ICCV.2017.212
\bibitem{article9}Dai J, Li Y, He K, et al. R-FCN: Object Detection via Region-based Fully Convolutional Networks. Advances in Neural Information Processing Systems. Curran Associates Inc.  2016. https://doi.org/10.1109/ICPICS50287.2020.9202049
\bibitem{article10}Tsung-Yi Lin, Piotr Doll'ar, Ross Girshick, Kaiming He,Bharath Hariharan, and Serge Belongie. Feature pyramidnetworks for object detection. InProc. IEEE Conf. Comp.Vis. Patt. Recogn., pages 2117–2125, 2017. https://doi.org/10.1109/CVPR.2017.106
\bibitem{article11}11.	Cai Z, Vasconcelos N. Cascade R-CNN: Delving into High Quality Object Detection.  2017.  https://doi.org /10.1109/CVPR.2018.00644
\bibitem{article12}Pang J, Chen K, Shi J, et al. Libra R-CNN: Towards Balanced Learning for Object Detection. 2019 IEEE/CVF Conference on Computer Vision and Pattern Recognition (CVPR). IEEE, 2020. https://doi.org/10.1109/CVPR.2019.00091
\bibitem{article13}Li Y, Chen Y, Wang N, et al. Scale-Aware Trident Networks for Object Detection. 2019 IEEE/CVF International Conference on Computer Vision (ICCV), 2019. https://doi.org/10.1109/ICCV.2019.00615
\bibitem{article14}Huang L, Yang Y, Deng Y, et al. DenseBox: Unifying Landmark Localization with End to End Object Detection. Computer Science, 2015
\bibitem{article15}iahui Yu, Yuning Jiang, Zhangyang Wang, Zhimin Cao, andThomas Huang. Unitbox: An advanced object detection net-work. InProc. ACM Int. Conf. Multimedia, pages 516–520.ACM, 2016. https://doi.org/10.1145/2964284.2967274
\bibitem{article16}Law H, Deng J. Cornernet: Detecting objects as paired keypoints. Proceedings of the European Conference on Computer Vision, 2018: 734-750. https://doi.org/10.1007/s11263-019-01204-1
\bibitem{article17}Law H, Teng Y, Russakovsky O, et al. CornerNet-Lite: Efficient Keypoint Based Object Detection. 2019
\bibitem{article18}Xingyi Zhou, Jiacheng Zhuo, Philipp Krahenbuhl. Bottom-Up Object Detection by Grouping Extreme and Center Points. The IEEE Conference on Computer Vision and Pattern Recognition (CVPR), 2019:850-859. https://doi.org/10.1109/CVPR.2019.00094
\bibitem{article19}Duan K, Bai S, Xie L, et al. CenterNet: Keypoint Triplets for Object Detection. International Conference on Computer Vision, 2019:6568-6577
\bibitem{article20}Zhu C, He Y, Savvides M. Feature selective anchor-free module for single-shot object detection. Proceedings of the IEEE Conference on Computer Vision and Pattern Recognition, 2019: 840-849. https://doi.org/10.1109/CVPR.2019.00093
\bibitem{article21}Wang J, Chen K, Yang S, et al. Region Proposal by Guided Anchoring. 2019 IEEE/CVF Conference on Computer Vision and Pattern Recognition (CVPR). IEEE,2019.https://doi.org/10.1109/CVPR.2019.00094.
\bibitem{article22}Tian Z, Shen C, Chen H, et al. Fcos: Fully convolutional one-stage object detection. Proceedings of the IEEE International Conference on Computer Vision,2019:96279636,https://doi.org/10.1109/ICCV.2019.00972
\bibitem{article23}Kong T, Sun F, Liu H, et al. FoveaBox: Beyond Anchor-based Object Detector.  2019
\bibitem{article24}Cao Y, Xu J, Lin S, et al. GCNet: Non-local Networks Meet Squeeze-Excitation Networks and Beyond. arXiv, 2019
\bibitem{article25}Wang X, Girshick R, Gupta A, et al. Non-local Neural Networks. Proceedings of the IEEE Computer Society Conference on Computer Vision and Pattern Recognition, 2017:7794-7803
\bibitem{article26}Hu J, Shen L, et al. Squeeze-and-Excitation Networks. IEEE Transactions on Pattern Analysis and Machine Intelligence,2017,42(8):2011-2023. https://doi.org/10.1109/TPAMI.2019.2913372
\bibitem{article27}X Wang, S Zhang, Z Yu, L Feng and W Zhang. Scale-Equalizing Pyramid Convolution for Object Detection. IEEE/CVF Conference on Computer Vision and Pattern Recognition (CVPR), 2020,13356-13365. https://doi.org/10.1109/CVPR42600.2020.01337
\bibitem{article28}Lin T Y, Goyal P, Girshick R, et al. Focal loss for dense object detection. Proceedings of the IEEE International Conference on Computer Vision, 2017: 2980-2988, https://doi.org/10.1109/TPAMI.2018.2858826.
\bibitem{article29}Zheng Z, Wang P, Liu W, et al. Distance-IoU Loss: Faster and Better Learning for Bounding Box Regression, arXiv:1911.08287.
\end{thebibliography}

\end{document}